\documentclass[
]{ceurart}

\usepackage{listings}
\usepackage{float}
\floatstyle{ruled}
\newfloat{prompt}{thp}{lop}
\floatname{prompt}{Prompt}
	
\usepackage{soul}
\begin{document}

\copyrightyear{2022}
\copyrightclause{Copyright for this paper by its authors.
  Use permitted under Creative Commons License Attribution 4.0
  International (CC BY 4.0).}

\conference{
14th International Workshop on Bibliometric-enhanced Information Retrieval, 24--28 March, 2024, Glasgow, Scotland}

\title{AI Insights: A Case Study on Utilizing ChatGPT Intelligence for Research Paper Analysis}





\author[1]{Anjalee De Silva}[%
orcid=0009-0001-5881-4408,
email=anjalee98@gmail.com,
]
\cormark[1]
\fnmark[1]
\address[1]{University of Queensland, Brisbane, Australia}

\author[2,3]{Janaka L. Wijekoon}[%
orcid=0000-0001-6092-3366,
email=janaka.wijekoon@gmail.com,
]
\cormark[1]
\fnmark[1]
\address[2]{Victorian Institute of Technology, Adelaide, South Australia}
\address[3]{Keio University, Yokohama, Japan}

\author[4]{Rashini Liyanarachchi}[%
orcid=0000-0003-0228-6658,
email=rashinikavindya@gmail.com,
]
\cormark[1]
\fnmark[1]
\address[4]{University of New South Wales(UNSW),  Sydney, Australia}

\author[5]{Rrubaa Panchendrarajan}[%
orcid=0000-0002-1403-2236,
email=r.panchendrarajan@qmul.ac.uk
]
\cormark[1]
\fnmark[1]
\address[5]{Queen Mary University of London, United Kingdom}

\author[6]{Weranga Rajapaksha}[%
orcid=0000-0001-8361-884X,
email=weranga.rajapaksha_gedara@mymail.unisa.edu.au
]
\cormark[1]
\fnmark[1]
\address[6]{University of South Australia, Adelaide, Australia}

\cortext[1]{Corresponding author.}
\fntext[1]{Author names appear alphabetically, and all authors contributed equally.}

\begin{abstract}
This paper discusses the effectiveness of leveraging Chatbot: Generative Pre-trained Transformer (ChatGPT) versions 3.5 and 4 for analyzing research papers for effective writing of scientific literature surveys. The study selected the \textit{Application of Artificial Intelligence in Breast Cancer Treatment} as the research topic. Research papers related to this topic were collected from three major publication databases Google Scholar, Pubmed, and Scopus. ChatGPT models were used to identify the category, scope, and relevant information from the research papers for automatic identification of relevant papers related to Breast Cancer Treatment (BCT), organization of papers according to scope, and identification of key information for survey paper writing. Evaluations performed using ground truth data annotated using subject experts reveal, that GPT-4 achieves 77.3\% accuracy in identifying the research paper categories and 
50\% of the papers were correctly identified by GPT-4 for their scopes. Further, the results demonstrate that GPT-4 can generate reasons for its decisions with an average of 27\% new words, and 67\% of the reasons given by the model were completely agreeable to the subject experts.
\end{abstract}


\maketitle

\section{Introduction}
Artificial Intelligence (AI), the term coined by John McCarthy in 1956 \cite{2} and discussed by prominent scientists such as Nikola Tesla in 1890 \cite{3}, Vannevar Bush in 1945 \cite{4}, and Alan Turing in 1950 \cite{5}, is a concept of machines thinking and applying knowledge. The ``Imitation Game" is a significant historical example of machines demonstrating the ability to think and use knowledge to solve problems \cite{5}. Ever since then, the research in AI has seen tremendous developments including the introduction of Neural Networks in 1990 \cite{6}. Notably, the momentum in AI was gained in various industries since 2017 \cite{9}, following the introduction of the transformer model: a parallel multi-head attention mechanism, by Ashish Vaswani et al. \cite{1}. The Transformer model led to the development of a Chatbot: Generative Pre-trained Transformer 3.5 (ChatGPT-3.5) in late 2022 \cite{7}, marking a revolutionary change in the AI domain \cite{34}. Subsequently, various industries have employed \cite{8,9,12,13,16,20}, debated \cite{11,14}, and disputed \cite{10,15} the use of GPT models.

Background studies revealed some studies embracing GPT \cite{18, 19, 20, 28} and focused on using GPT to automate academic writing \cite{23, 25, 29}. Whereas, some approached GPT cautiously \cite{21, 22, 26, 27} and some emphasised the need for regulation and introducing new guidelines for using generative AI \cite{24}. Consequently, as a part of our ongoing research project, this paper discusses the effectiveness of using GPT models to analyze research papers for writing scientific literature surveys.

The project is carried out to produce a review paper revealing how AI applications are used in Breast Cancer Treatment (BCT). The overall study comprises five stages (Refer to Figure \ref{fig:methodology}) starting from the construction of a taxonomy depicting the branches of BCT, followed by the research paper collection, and automatic analysis of research papers for drafting the survey paper. Subsequently, this paper presents the effectiveness of using ChatGPT to automatically analyze the gathered research papers.

We used both GPT-3.5 (2022 January update) and GPT-4 (2023 April update) models to automatically analyze the research papers. Various information presented in the research paper including the title, abstract, and textual content was used at different stages of the study. Research papers gathered from three major publication databases Google Scholar, Pubmed, and Scopus were merged without duplicates to form a unified corpus related to \textit{AI in BCT}. ChatGPT models were employed in this corpus to identify the research paper categories and scope automatically and to retrieve information required for survey paper writing. The performance of the models in identifying the category and scope of a research paper was evaluated against ground truth data annotated by subject experts. Further, the capability of the models to generate reasons for their own decisions was analyzed with the help of the same subject experts. Results of our experiments are presented in Section \ref{sec:Results and Discussion}. 

\section{Methodology}

\begin{figure}[!t]
  \centering
  \includegraphics[width=0.7\textwidth]{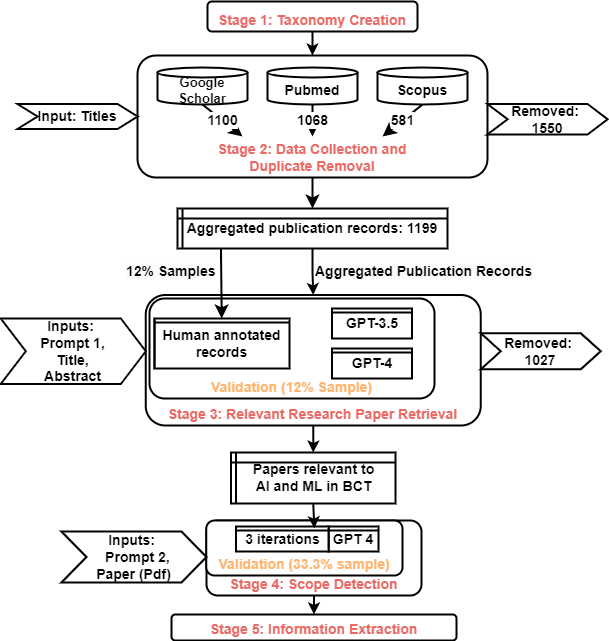}
  \caption{Overall Methodology}
  \label{fig:methodology}
\end{figure}

Figure \ref{fig:methodology} illustrates the flow of the methodology carried out in this study via various stages. Stage 1 comprises the construction of a taxonomy depicting the branches of BCT. Following this, in stage 2, the search queries were developed based on the taxonomy, and the research papers were gathered by querying Google Scholar, Pubmed, and Scopus, followed by duplication removal to form a unified corpus of research articles related to \textit{AI in BCT} \footnote{Automating scripts can be found at https://github.com/janakawest/ScholrlyDatagathering}.

Stage 3 focused on identifying the research paper category to filter out the relevant papers related to BCT. We compared the performance of both GPT-3.5 and GPT-4 in identifying research paper categories using an human-annotated sample of 12\% research papers from the unified corpus. Then, the best GPT model was used to identify research paper categories, and research papers related to \textit{AI in BCT} proceeded to the next stage. In Stage 4, we identified the BCT scope of the research papers to organize them in the survey paper according to their area of study. We also validated the performance of GPT model in identifying the scope of a research paper using a human-annotated sample of 33\% relevant research papers. Finally, in Stage 5, we extracted the relevant information from the research paper required to write the survey article. 

\subsection{Taxonomy Construction}\label{sec:Tax}
\begin{figure} [!t]
  \centering
  \includegraphics[width=0.7\textwidth]{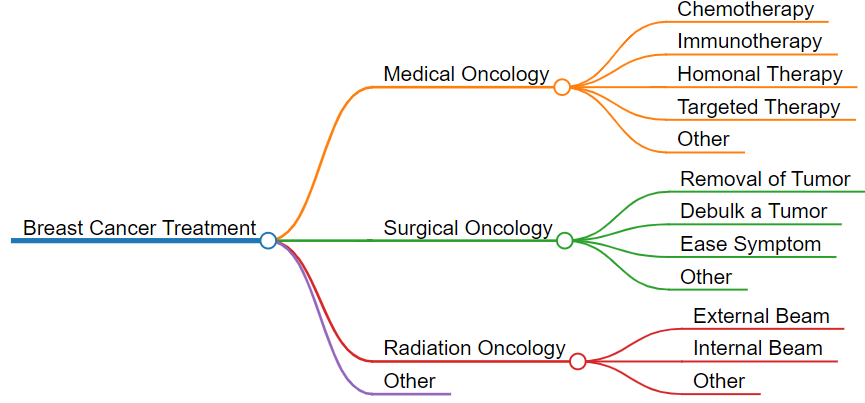}
  \caption{Taxonomy of \textit{BCT} used for the Study}
  \label{fig:taxonomy}
\end{figure}

Our study began with constructing a taxonomy depicting all the branches of BCT. This taxonomy was used to retrieve research papers from publication databases and to categorize and organize various treatment methods into a structured format for easier understanding and analysis. Figure \ref{fig:taxonomy} depicts the taxonomy used in this study.

Specifically, the BCT options were primarily organized under three broader oncological categories namely medical oncology, surgical oncology, and radiation oncology. Further, each category is divided into sub-categories. We included the type \textit{Other} in each branch to indicate the treatment types excluded in the taxonomy. For example, the \textit{Other} type under \textit{Medical Oncology} was included to cover all the sub-branches of medical oncology excluded in our taxonomy. As this is an ongoing project, we hope to extend our taxonomy to cover all the BCT types, and this is discussed as a future work in Section \ref{sec: Conclusion}.

\subsection{Data Collection}
\label{sec:DatCol}
We collected research articles from three major publication databases, Google Scholar \footnote{https://scholar.google.com/}, Pubmed \footnote{https://pubmed.ncbi.nlm.nih.gov/}, and Scopus \footnote{https://www.scopus.com/}. Each database was queried independently using the keywords related to the topic \textit{AI for BCT}. The search keywords were defined based on the taxonomy (Refer Figure \ref{fig:taxonomy}), such that each keyword represents a node or a branch of the taxonomy. For example, the search keyword \textit{Radiology} represents all three sub-branches of \textit{Radiation Oncology}, whereas the search keyword \textit{Chemotherapy} represents a node in the taxonomy. Together with the keywords, the search queries included \textit{``AI'', ``Artificial Intelligence'', ``Breast cancer'', `` <<breast cancer treatment type>>''} to target the search to BCT. For example, \textit{ ``AI'', ``Artificial Intelligence'', ``Breast cancer'', ``Radiology''} is an example search query used to retrieve research papers related to radiology in BCT. Altogether, we used 13 keywords resulting in 13 search queries, and the research papers were retrieved by querying the publication database using these search queries. 

\begin{figure}[!t]
  \centering
  \includegraphics[width=0.7\textwidth]{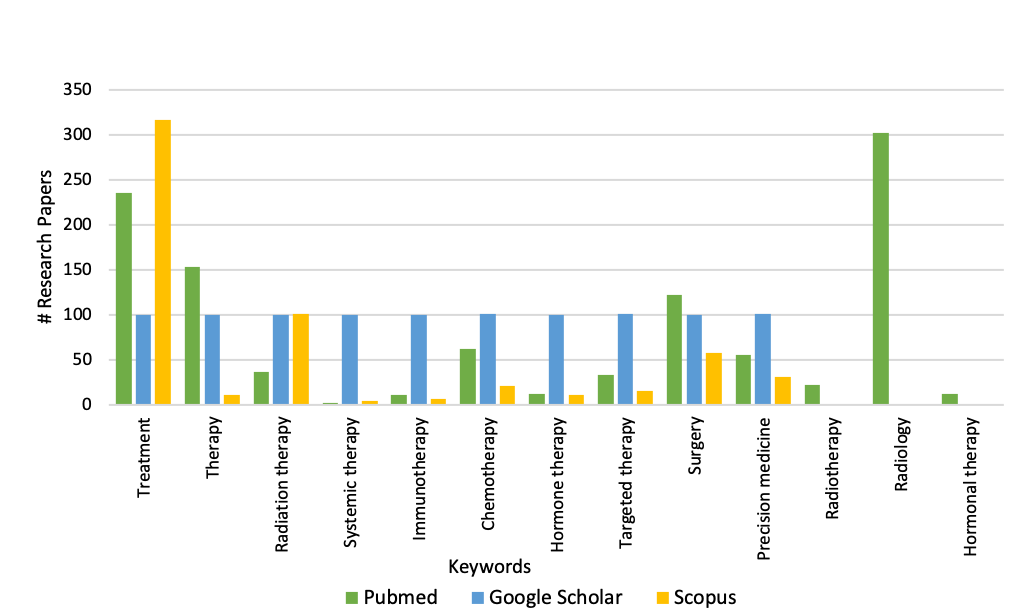}
  \caption{Distribution of Research Papers Retrieved using Keyword related to \textit{BCT}}
  \label{fig:keyword stat}
\end{figure}

We used the built-in API of Google Scholar and Pubmed to query the research papers automatically, and Scopus was queried using the user interface manually. For Google Scholar API, we had to mention how many publication records the scraper should retrieve manually. For this study, we limited each query to get 110 records per search to avoid the increase in noise. Figure \ref{fig:keyword stat} presents the number of research papers retrieved for each source for the 13 search keywords. 

Once the research papers were collected, we removed the duplicate papers within the source as well as across the sources to generate a unified corpus of research papers related to the topic \textit{AI in BCT}. We used the title of the research papers to identify the duplicates in this stage of the study. Table \ref{tab:dataset stat} shows the statistics of the unified corpus. Referring to the table, it can be observed, that Pubmed and Google Scholar APIs resulted in a higher number of duplicates (roughly 50\% of duplicate records) due to their nature in indexing research papers from multiple other repositories and their retrieval mechanism. After removing the duplicate research articles within each source, we ended up with 462 - 516 records per each source. Further, we merged the three repositories and removed the duplicates across the sources resulting in a final corpus of size 1199. Figure \ref{fig:methodology} illustrates the number of records removed as duplicates.

\begin{table*}[]
\caption{Statistics of the Dataset}
  \label{tab:dataset stat}
\small
\begin{tabular}{p{6cm}p{3.5cm}p{3.5cm}}
\toprule
Source         & \# of Papers Collected & \# of Unique Papers    \\ \midrule
Pubmed         & 1,068                      & 516                                  \\ 
Google Scholar & 1,100                      & 511                                   \\ 
Scopus         & 581                      & 462                                   \\ \midrule
\# of Unique Papers across Sources    &    & \textbf {1,199}  \\
\bottomrule
\end{tabular}
\end{table*}

\subsection{Relevant Research Paper Retrieval}\label{sec:relevant-paper-retrieval}
The research papers retrieved through the search keywords in stage 1 did not ensure the retrieval of relevant papers related to \textit{AI in BCT}. A manual analysis of the unified corpus revealed that, in addition to the papers related to AI in BCT, it comprised a mix of research papers related to AI for breast cancer diagnosis, some review papers in breast cancer research, research papers focusing on both BCT and diagnosis, along with completely irrelevant papers. Therefore, this step focused on automatically identifying the research paper categories to filter out articles related to BCT from the unified corpus. 

Given the title, and abstract of the research paper, the ChatGPT model was instructed to identify the objective of the paper from the aforementioned options. For this purpose, we designed the Prompt \ref{prompt1} as a single-option multiple-choice question, allowing the model to choose the right answer from the given options. 

\begin{prompt}[!t]
\begin{footnotesize}
\begin{verbatim}
Title:
Abstract:
Analyze the scientific paper's title and abstract to determine the
most accurate description of its objectives
from the following options:

A: A Review or survey paper summarizing research related to breast cancer
B: A Research study on breast cancer diagnosis
C: A Research study on breast cancer treatment
D: A Research study on breast cancer diagnosis and treatment
E: None of the above
F: Not sure

Share your response in the format:
Answer
Reason
\end{verbatim}
\end{footnotesize}
  \caption{Category Identification Prompt}
  \label{prompt1}
\end{prompt}

We included option E indicating the research articles that are not related to BCT, and option F was included in Prompt \ref{prompt1} to enable the model to indicate that it is unsure about the decision. We noticed that the model may choose different options for the same article's title and abstract during different prompt executions as it is a generative model \cite{32}. Therefore, for each GPT model, we executed Prompt \ref{prompt1} three times and chose the majority option selected by the model. If no majority was found, then we marked the ChatGPT answer as \textit{F: Not sure}, to indicate the confusion of the model in determining the answer. Once the majority option was selected, papers indicated as either option \textit{C} or \textit{D} were filtered out as relevant papers for the next stage of the study. We ended up with 143 relevant papers focusing on BCT as their key objective. 

We used both GPT-3.5 and GPT-4 for this stage of the study and the performance of both models on categorizing the research papers (Refer Section \ref{sec:evaluation-category detection}) was evaluated using 12\% {human-annotated} sample of the unified corpus. Due to the unsurpassed performance of the GPT-4 model, compared to GPT-3.5, we carried out the remaining study using the GPT-4 model. 

\subsection{Scope Detection} \label{sec:scope-detection}
Each relevant paper focusing on BCT may discuss the application of AI in different BCT. Therefore, the next stage of our study was focused on automatically identifying the \textit{scope} of the research paper, indicating the BCT type(s) experimented by the authors. In this stage, we defined the \textit{scope} as a path in the taxonomy (Figure \ref{fig:taxonomy}). For example, \textit{Medical Oncology - Chemotherapy} is a possible scope of BCT. Similar to the relevant paper detection, we designed Prompt \ref{prompt2} for ChatGPT as a multiple-choice question with all the possible scopes (all the paths in the taxonomy) as the answers. This resulted in 15 possible answers for the scope detection prompt. Here, options A to M indicate the paths in the taxonomy. 

\begin{prompt}
\begin{footnotesize}
\begin{verbatim}
Make sure you thoroughly read through the entire research paper and classify the content of
the paper from the following options:

A: Chemotherapy of Medical Oncology
B: Immunotherapy of Medical Oncology
C: Hormonal therapy of Medical Oncology
...
L: Other treatments of Radiation Oncology
M: Other treatment categories directly related to breast cancer 
N: Reviews or meta-analyses on breast cancer treatments
O: Studies not directly related to the treatment of breast cancer 

In case, if the authors contributed on many areas of treatments, select all the necessary 
options from the list A to M as the answer. Else, If the paper provides a detailed review of
existing breast cancer treatments, then choose Option "N" as the answer. Otherwise, If the
content of the paper does NOT contribute to breast cancer treatment at all, then choose 
option "O" as the answer. 

PLEASE Do not provide any explanations. Just mention the suitable options. 

NOTE: Make sure you read the entire research paper to select the options.
\end{verbatim}
\end{footnotesize}
  \caption{Scope Detection Prompt}
  \label{prompt2}
\end{prompt}

We executed Prompt \ref{prompt2} by uploading the PDF file of each relevant paper to GPT-4. The model was instructed to read the entire research article and identify the main focus of the research paper from the list of scope options. The model was instructed to select more than one answer among options A-M if the authors had contributed to more than one BCT type. We ran the prompt three times for each research paper, and the options that appeared in more than one execution were chosen as the scope of the research paper.

Similar to the previous stage, we evaluated the performance of the GPT-4 model in correctly identifying the scope of the research paper using a human-annotated sample of size 33\% of the relevant papers. Further details on this experiment are discussed in Section \ref{sec:evaluation-scope-detection}.

\subsection{Information extraction}
After utilizing GPT-4 to identify the scope of the paper, this stage of the study was aimed at extracting the information required for survey paper writing. We uploaded the paper to GPT-4 and provided Prompt \ref{prompt3} to extract specific details, including background \& objective, methods, key findings, conclusions, and limitations of the study. Subject experts verified the extracted information for a sample research paper\footnote{Data of a large sample is not presented due to page limitation } \cite{31} and the observations are discussed in Section \ref{sec:results-information-extraction}. As this is ongoing research, Prompt \ref{prompt3} will be further enhanced to retrieve scope-specific information as one of the main future works.

\begin{prompt}
\begin{footnotesize}
\begin{verbatim}
Analyze the research article thoroughly and generate a concise summary in tabular form that
captures the fundamental aspects of the study. Include the following key points:

1. Background and Objective:
   - Context: Briefly describe the background and context of the research.
   - Objective: Specify the goals or discoveries the authors aimed to achieve.
2. Methods:
   - Methodologies: Summarize the methods, experiments, or analysis employed by the authors
   in conducting their research.
3. Key Findings:
   - Main Results: Provide a clear overview of the primary results or discoveries presented
   in the paper.
4. Conclusion:
   - Final Takeaways: Summarize the concluding remarks of the research.
   - Future Directions: Highlight any recommendations or future directions suggested by 
   the authors.
5. Limitations:
   - Constraints: Identify and summarize any limitations or constraints mentioned in the study.

Ensure that the summary is comprehensive, and coherent, and avoids unnecessary jargon. 
Present the information in a well-organized tabular format for easy comprehension.
\end{verbatim}
\end{footnotesize}
  \caption{Information Extraction Prompt}
  \label{prompt3}
\end{prompt}

\section{Results and Discussion}
\label{sec:Results and Discussion}
In this section, we analyze and present the performance of GPT in 1.) identifying research paper categories using the title and abstract, 2.) analyzing the reasons provided by the GPT for category detection, and 3.) identifying the scope of a research paper by reading the entire paper. The temperature of both GPT models was not altered assuming their default temperature values, i.e., 1 for both models \cite{35}.

\subsection{Relevant Research Paper Retrieval}
We randomly sampled around 12\% of the research papers from the initial corpus of size 1199. However, there were some research papers in the sample without abstracts and we had to drop those papers for a fair evaluation of the model in identifying the research paper category using title and abstract. This resulted in a final sample size of 132. The sampled data were used to analyze the performance of GPT-3.5 and GPT-4, and the evaluation process and the performance of the model are discussed in the following sections.   

\subsubsection{Performance on Category Identification}\label{sec:evaluation-category detection}
The selected sample papers were first given to two subject experts for annotation based on the title and abstract, and the human-annotated data were used to evaluate the performance of the model in identifying the research paper category. The subject experts annotated the sample using the six options listed in Section \ref{sec:relevant-paper-retrieval}, and any disagreements between the subject experts were resolved through discussion. The accuracy of the models on the sample data is presented in Table \ref{tab:performance-relevancy-classification}. The table indicates the average accuracy of three iterations and the accuracy of the majority option selection by the models during the three iterations (i.e., two selections out of three iterations). It is noteworthy that, in both cases, the GPT-4 model significantly outperformed the GPT-3.5 model in identifying research paper categories.

\begin{table}[]
\caption{Performance of ChatGPT Models in Category Identifation}
  \label{tab:performance-relevancy-classification}
\small
\begin{tabular}{lllll}
\toprule
Model         & Accur. (Avg.) & Accur. (Maj.) & Avg. Response Length  & New Words Response (Avg.\%)  \\ \midrule
GPT-3.5         & 65.15\% ($\pm$6.8)&65.15\%& 45& 18\% \\ 
GPT-4            & \textbf{76.21\%} ($\pm$5) &\textbf{77.3\%} & 68.5& 27\% \\ \bottomrule
\end{tabular}
\end{table}

\begin{figure}
  \centering
  \includegraphics[width=0.5\textwidth]{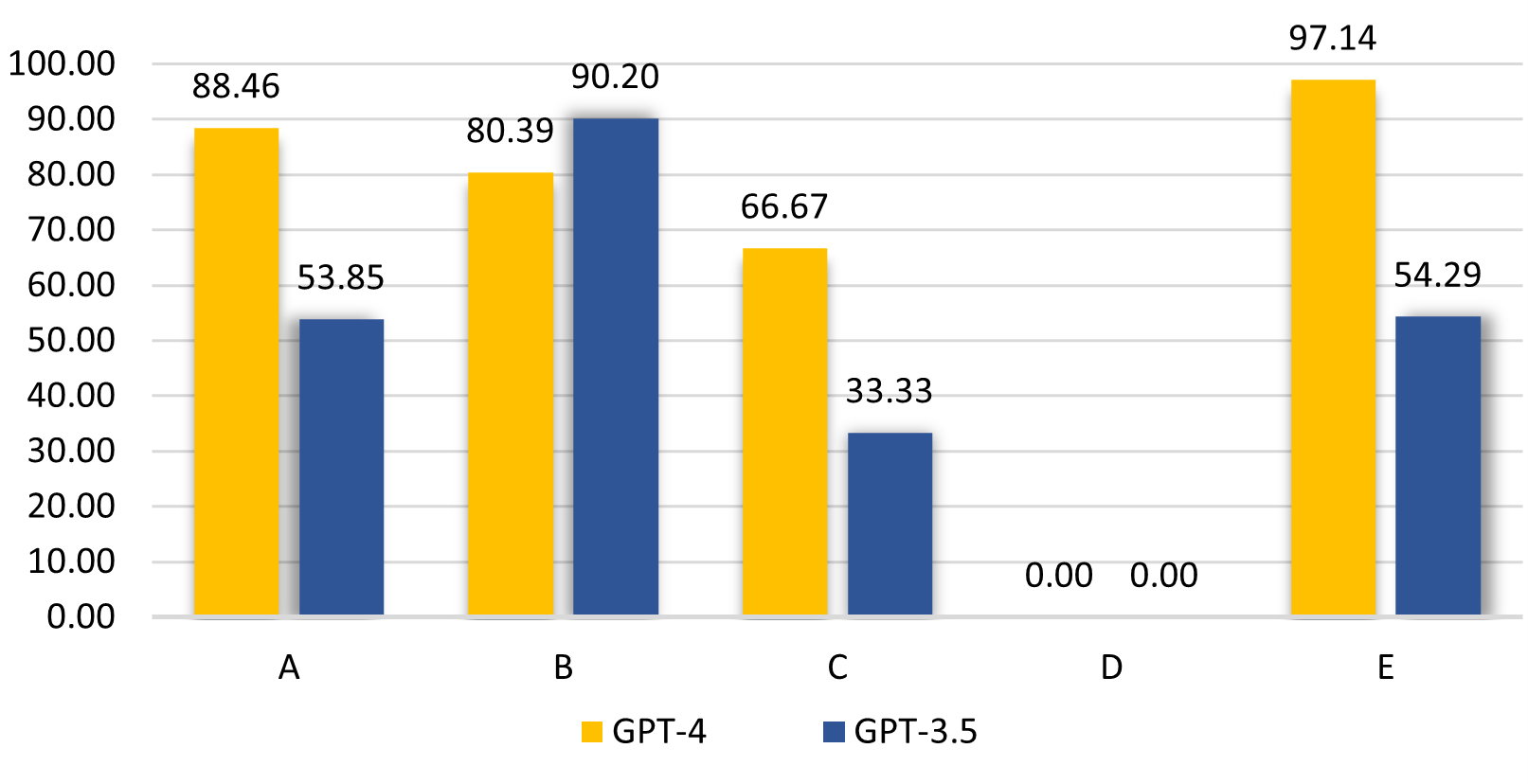}
  \caption{Category-wise Accuracy of ChatGPT Models (Refer to Prompt \ref{prompt1} for Categories)}
  \label{fig:category-based-performance}
\end{figure}

\begin{figure}
  \centering
  \includegraphics[width=.6\textwidth]{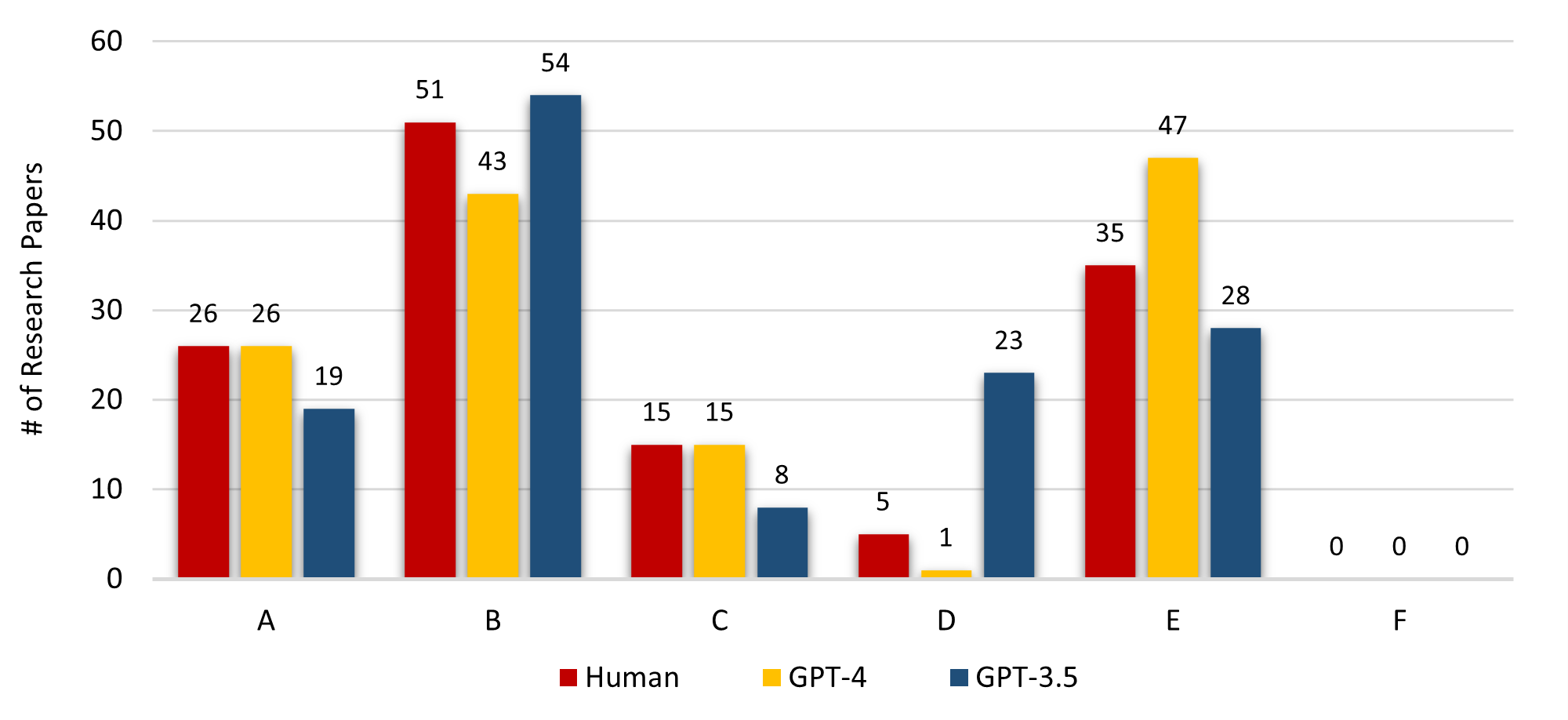}
  \caption{Statistics of the Research Papers Categorized by Human and ChatGPT Models (Refer to Prompt \ref{prompt1} for Categories)}
  \label{fig:Ground-truth-research-paper-statistics}
\end{figure}

Furthermore, we analysed the accuracy of identifying individual research paper categories. Figures \ref{fig:category-based-performance} and \ref{fig:Ground-truth-research-paper-statistics} show the category-wise performance of the models and the number of research articles classified under each category by subject experts Vs. ChatGPT models. It can be observed that the performance of the GPT-4 model is significantly higher than GPT-3.5 across all the categories, except the research articles focusing on breast cancer diagnosis (\textit{Option B}). This could be possibly due to the GPT-4 model classifying some of the breast cancer detection papers as \textit{Option E - None of the above}. Further, both models show drastically different behavior in identifying research articles focusing on both diagnosis and treatment (\textit{Option D}) and fail to correctly identify any of the papers belonging to this category. Interestingly, none of the models choose the \textit{Option F - Not sure} within the sample data. 

\subsubsection{Reasoning Analysis}
\begin{figure}
  \centering
  \includegraphics[width=0.55\textwidth]{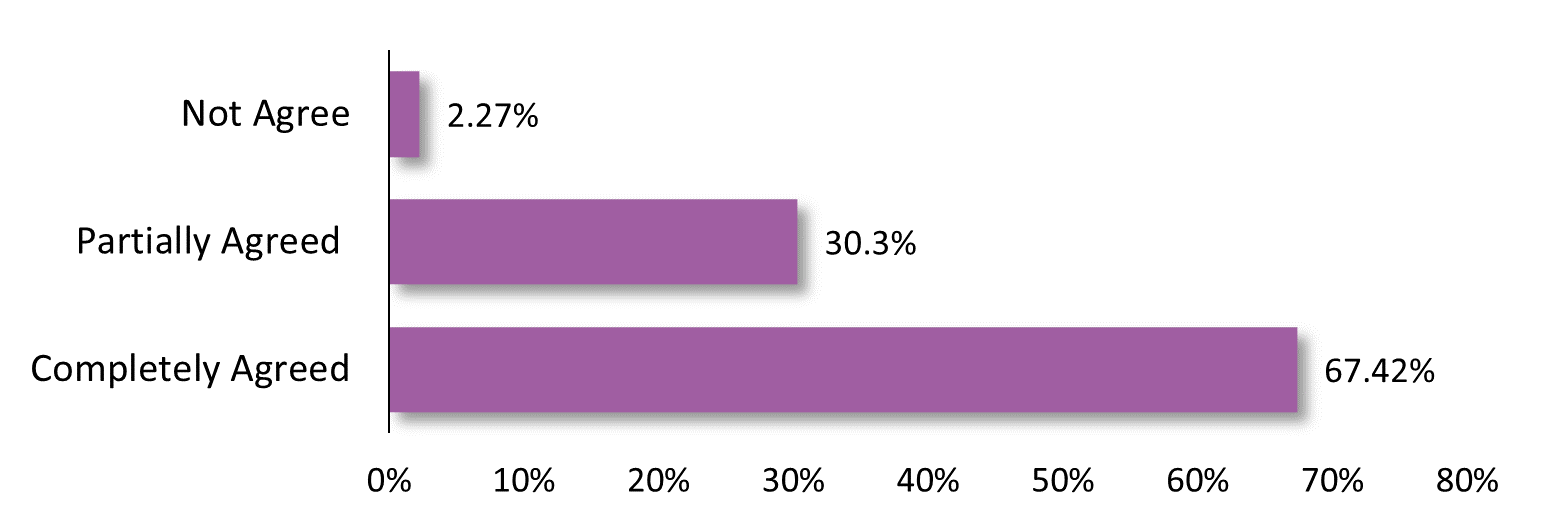}
  \caption{Distribution of Agreement Level of Reasoning}
  \label{fig:Resoning-agreement}
\end{figure}

In addition to the capability of the GPT model to identify the category of a research paper, we analyzed their reasoning capability while making a decision \cite{33}. We instructed the GPT model to give a reason for choosing the option among the six categories in Prompt \ref{prompt1}. Our domain experts read the research article's title, abstract, and GPT's choice and the reasons generated by the models, and marked it as either \textit{Completely Agreed}, \textit{Partially Agreed},  or \textit{Not Agree} indicating the level of accurate reasoning the model produces with respect to its choice. Figure \ref{fig:Resoning-agreement} presents the distribution of agreement level of the reasoning of the GPT-4 model. It is evident that the model produces completely agreeable reasoning most of the time (67.42\%), and very rarely it generates invalid reasoning (2.27\%). Table \ref{tab:performance-relevancy-classification} shows the average length of the reasons generated by both GPT-3.5 and GPT-4 models. It can be observed, that GPT-3.5 generates relatively shorter reasons compared to GPT-4.   

\begin{figure}
  \centering
  \includegraphics[width=0.8\textwidth]{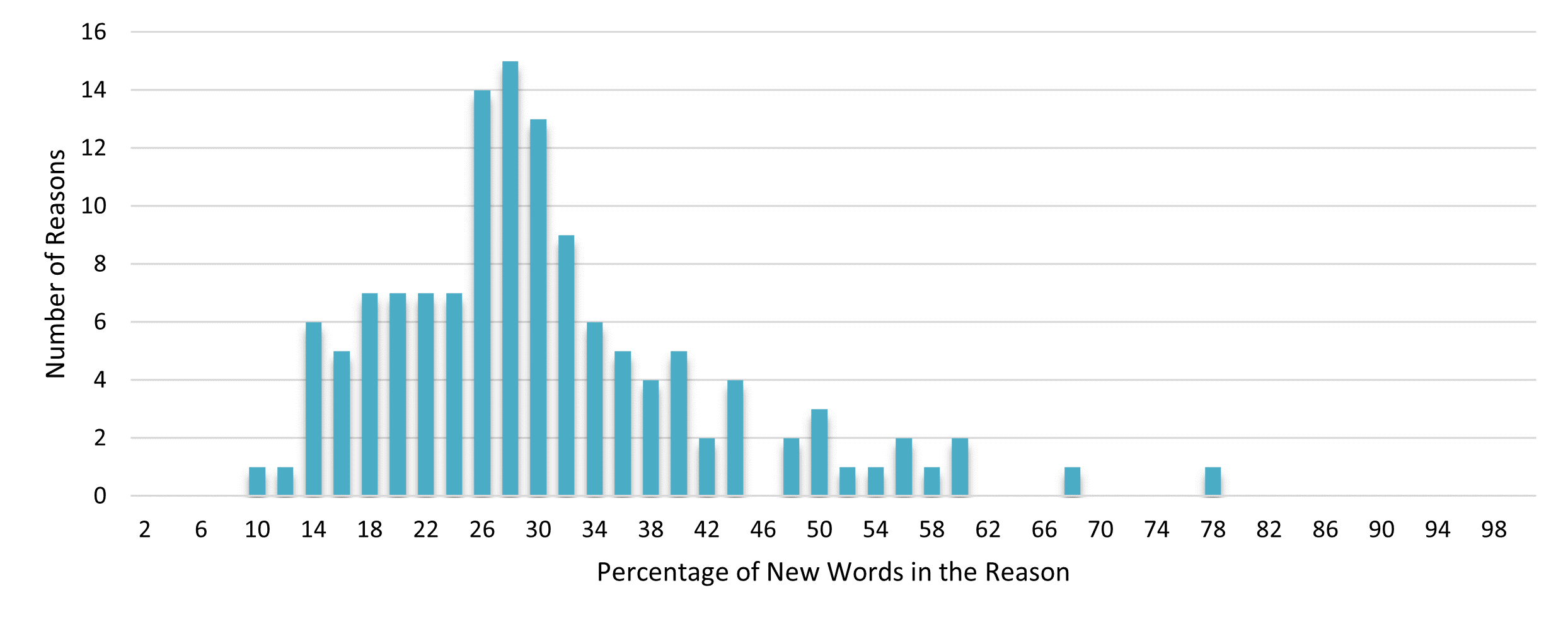}
  \caption{Distribution of Percentage of New Words Used by GPT-4 for Reasoning}
  \label{fig:New-words-distributions}
\end{figure}

We further analyze the capability of the model to generate reasons with its wording instead of copying the content from the abstract and the title. We computed the percentage of new words appearing in the reason compared to the abstract. Here, the stop words\footnote{https://www.nltk.org/search.html?q=stopwords} were not considered as words. The observed results are presented in Table \ref{tab:performance-relevancy-classification}. The table presents the average number of new words produced by both models, and interestingly GPT-4 model produces relatively more new words in the reason compared to GPT-3.5, though the former generates lengthier reasons. The distribution of the percentage of new words generated by the GPT-4 model during reasoning is presented in Figure \ref{fig:New-words-distributions}. It can be noticed that the GPT-4 model generates 25-30\% of new words when reasoning and the highest value of 78\% is also observed in the sample data.   

\subsection{Scope Detection}\label{sec:evaluation-scope-detection}
To effectively evaluate the model's ability to identify the scope of selected papers in BCT, a detailed annotation process was employed using 15 predefined scope options mentioned in Prompt \ref{prompt2}. Similar to the category identification, initially, two subject experts were presented with the full text of approximately 33\% of the relevant papers related to BCT and asked to independently determine its treatment scope from the above-mentioned options. Noting that a single paper could potentially align with multiple scopes. In instances of disagreement between the annotators, resolution was achieved through discussion. Subsequently, GPT-4 was utilized to process each paper, providing both Prompt \ref{prompt2} and the PDF file over three separate iterations for thorough analysis.

The yielded results are presented in Table \ref{table:scope detection}. Some of the papers have multiple scopes whereas some of the papers focus on a single scope. GPT-4 correctly identified the scope of 50\% of the research papers, and for 22\%, it identified a portion of the scope annotated by subject experts. Within the 22\% intermediate matched, there were 4\% instances where GPT identified a smaller scope than the experts and 16\% times it provided a broader range of scopes compared to what subject experts had identified.

\begin{table}[!t]
\caption{Scope Detection Percentage Distribution of GPT-4}
\small
\begin{tabular}
{ll}
\toprule
Match Level & Percentage\\ \midrule
Complete Match      &     50\%        \\
Intermediate Match       &     22\%      \\
No Match            &     28\%     \\ \bottomrule
\end{tabular}
\label{table:scope detection}
\end{table}

\subsection{Information extraction}\label{sec:results-information-extraction}
This stage of the research is still under development, hence we present only the information retrieved by GPT-4 for a sample research paper \cite{31} for the information extraction Prompt \ref{prompt3}. The extracted information is presented in Table \ref{Table:Information extraction}. According to the table, the GPT-4 successfully extracted the important information. 

\begin{table}[!t]
\centering
\caption{Information Extracted from a Sample Research Paper \cite{31}}
\footnotesize
\begin{tabular}{|p{2cm}|p{11.5cm}|}
\hline
\textbf{Aspect} & \textbf{Details} \\ \hline
Background and Objective & \textbf{Context:} Focuses on improving radiotherapy planning through automated contouring tools. \newline \textbf{Objective:} Assessing the performance of AI-Rad, a machine-learning automated contouring tool, compared to manual and another automated tool. \\ \hline
Methods & \textbf{Methodologies:} AI-Rad used for automated contouring in 28 patients, compared with manual and SS contours. Evaluation metrics included Dice similarity coefficient, sensitivity, precision, and Hausdorff distance. \\ \hline
Key Findings & \textbf{Main Results:} AI-Rad produced clinically acceptable contours, often superior to SS, with higher efficiency and minimal editing requirements. Some structures, like the larynx, were challenging, but AI-Rad showed promise in improving radiotherapy planning efficiency. \\ \hline
\end{tabular}
\label{Table:Information extraction}
\end{table}

\subsection{Limitations}
This section details the limitations we encountered during the entire process of our study. 
\begin{itemize}
    \item \textbf{Noisy data retrieval} - Given the nature of Google Scholar and Pubmed, they retrieve and compile data from various publication databases. Consequently, when we automatically queried using APIs, the search results had several flaws, including missing abstracts, incomplete titles, abstracts and titles extracted from the body of texts, partial abstracts, or a high noise level in the retrieved data. For instance, when utilizing the scholarly API\footnote{https://pypi.org/project/scholarly/}, it was observed that requesting large numbers, such as 200, led to a significant portion of these results containing irrelevant or incorrect data. Therefore, we limited the Google Scholar search to the first 110 research papers.
    \item \textbf{Inconsistent Chat-GPT response} - As we already mentioned, Both GPT models were inconsistent in generating the response and provided different options in various iterations. Furthermore, when executing the prompt in bulk (e.g. 5 research papers together), we received responses for more than the number of research articles queried; maybe GPT is now lazy as mentioned in \cite{17}. 
    \item \textbf{Limited Chat-GPT functionality} - We noticed that the inherent performance limitations of GPT models, specifically when using GPT-4, the message limit of 50 messages for every 3 hours slow down our analysis significantly \cite{30}. Moreover, even with the paid version, additional charges for API querying limited us to manual execution of all the prompts hindering the efficiency of the automation.
    \item \textbf{Iterative prompt creation} - One observation we made during the prompt creation was that the prompts had to undergo several iterations of edits to refine them and achieve the optimal results reported in this study. We included only the prompts that achieved optimal performance due to space limitations.

\end{itemize}

\section{Conclusion}\label{sec: Conclusion}
This paper presents a preliminary study of ongoing research focusing a survey on the topic \textit{AI for breast cancer treatment (BCT)}. We analyzed the effectiveness of using ChatGPT models for automating the research paper analysis for survey paper writing. Specifically, we evaluated GPT-3.5 and GPT-4 for automatic paper category identification, scope detection, and information extraction. Experiment results compared to ground truth data reveal GPT-4 model can be used to identify the category of the research papers for automating the research paper analysis. However, the model seems to struggle when accurately identifying the scope of a research study. Further, we detailed the limitations that could potentially hinder the adoption of GPT models for scholarly work. 
As a future work, we will extend this work to come up with a comprehensive taxonomy of BCT and compile a survey article on \textit{AI for BCT}. 
\bibliography{ceur}
\end{document}